\documentclass[letterpaper]{article} % DO NOT CHANGE THIS
\usepackage{aaai24}  % DO NOT CHANGE THIS
\usepackage{times}  % DO NOT CHANGE THIS
\pdfoutput=1
\usepackage{helvet}  % DO NOT CHANGE THIS
\usepackage{courier}  % DO NOT CHANGE THIS
\usepackage[hyphens]{url}  % DO NOT CHANGE THIS
\usepackage{graphicx} % DO NOT CHANGE THIS
\usepackage{natbib}  % DO NOT CHANGE THIS AND DO NOT ADD ANY OPTIONS TO IT
\usepackage{caption} % DO NOT CHANGE THIS AND DO NOT ADD ANY OPTIONS TO IT
\usepackage{algorithm}
\usepackage{algorithmic}
\usepackage{amsmath,amsfonts}
\usepackage{amssymb}
\usepackage{bbding}
\usepackage{booktabs}
\usepackage{multirow}
\usepackage{subfig}
\usepackage{lineno}
\usepackage{cite}
\usepackage{color}
\usepackage{newfloat}
\usepackage{listings}
\DeclareCaptionStyle{ruled}{labelfont=normalfont,labelsep=colon,strut=off} % DO NOT CHANGE THIS
\floatstyle{ruled}
\newfloat{listing}{tb}{lst}{}
\floatname{listing}{Listing}

\title{TCI-Former: Thermal Conduction-Inspired Transformer for Infrared Small Target Detection}
\author{
    Tianxiang Chen\textsuperscript{\rm 1,2,3}\thanks{Work done during an internship at Alibaba Group.},
    Zhentao Tan\textsuperscript{\rm 1,2,3},
    Qi Chu\textsuperscript{\rm 1,3}\thanks{Qi Chu is the corresponding author.},
    Yue Wu\textsuperscript{\rm 2},
    Bin Liu\textsuperscript{\rm 1,3},
    Nenghai Yu\textsuperscript{\rm 1,3}
}
\affiliations{
    \textsuperscript{\rm 1}School of Cyber Science and Technology, University of Science and Technology of China\\
    \textsuperscript{\rm 2}Alibaba Group\\
    \textsuperscript{\rm 3}Key Laboratory of Electromagnetic Space Information, Chinese Academy of Sciences\\
    \{txchen,tzt\}@mail.ustc.edu.cn, matthew.wy@alibaba-inc.com, \{qchu, flowice, ynh\}@ustc.edu.cn\\
}

\usepackage{bibentry}
\begin{document}

\maketitle

\begin{abstract}
Infrared small target detection (ISTD) is critical to national security and has been extensively applied in military areas. ISTD aims to segment small target pixels from background. Most ISTD networks focus on designing feature extraction blocks or feature fusion modules, but rarely describe the ISTD process from the feature map evolution perspective. In the ISTD process, the network attention gradually shifts towards target areas. We abstract this process as the directional movement of feature map pixels to target areas through convolution, pooling and interactions with surrounding pixels, which can be analogous to the movement of thermal particles constrained by surrounding variables and particles. In light of this analogy, we propose Thermal Conduction-Inspired Transformer (TCI-Former) based on the theoretical principles of thermal conduction. According to thermal conduction differential equation in heat dynamics, we derive the pixel movement differential equation (PMDE) in the image domain and further develop two modules: Thermal Conduction-Inspired Attention (TCIA) and Thermal Conduction Boundary Module (TCBM). TCIA incorporates finite difference method with PMDE to reach a numerical approximation so that target body features can be extracted. To further remove errors in boundary areas, TCBM is designed and supervised by boundary masks to refine target body features with fine boundary details. Experiments on IRSTD-1k and NUAA-SIRST demonstrate the superiority of our method.
\end{abstract}

\section{Introduction}
Infrared small target detection (ISTD) is challenging because targets are so small that may easily get ignored by generic segmentation networks. Besides, infrared images are of low contrast and low quality, which also bring challenges to this task. Since generic segmentation networks fail to perform well on this task, we hope to explore a new perspective and design a precise and explainable method for ISTD.

ISTD methods are generally categorized into traditional methods and deep-learning-based methods. In early stages, for lack of public ISTD dataset, researchers are limited to traditional methods \cite{sun2020infrared,marvasti2018flying,zhang2019infrared,han2019local}. However, these methods relying so much on prior knowledge and handcraft features that inevitably suffer very limited performances on images with characteristics inconsistent with the model assumptions. 

\begin{figure}
    \centering
    \includegraphics[width=0.45\textwidth,height=0.23\textwidth]{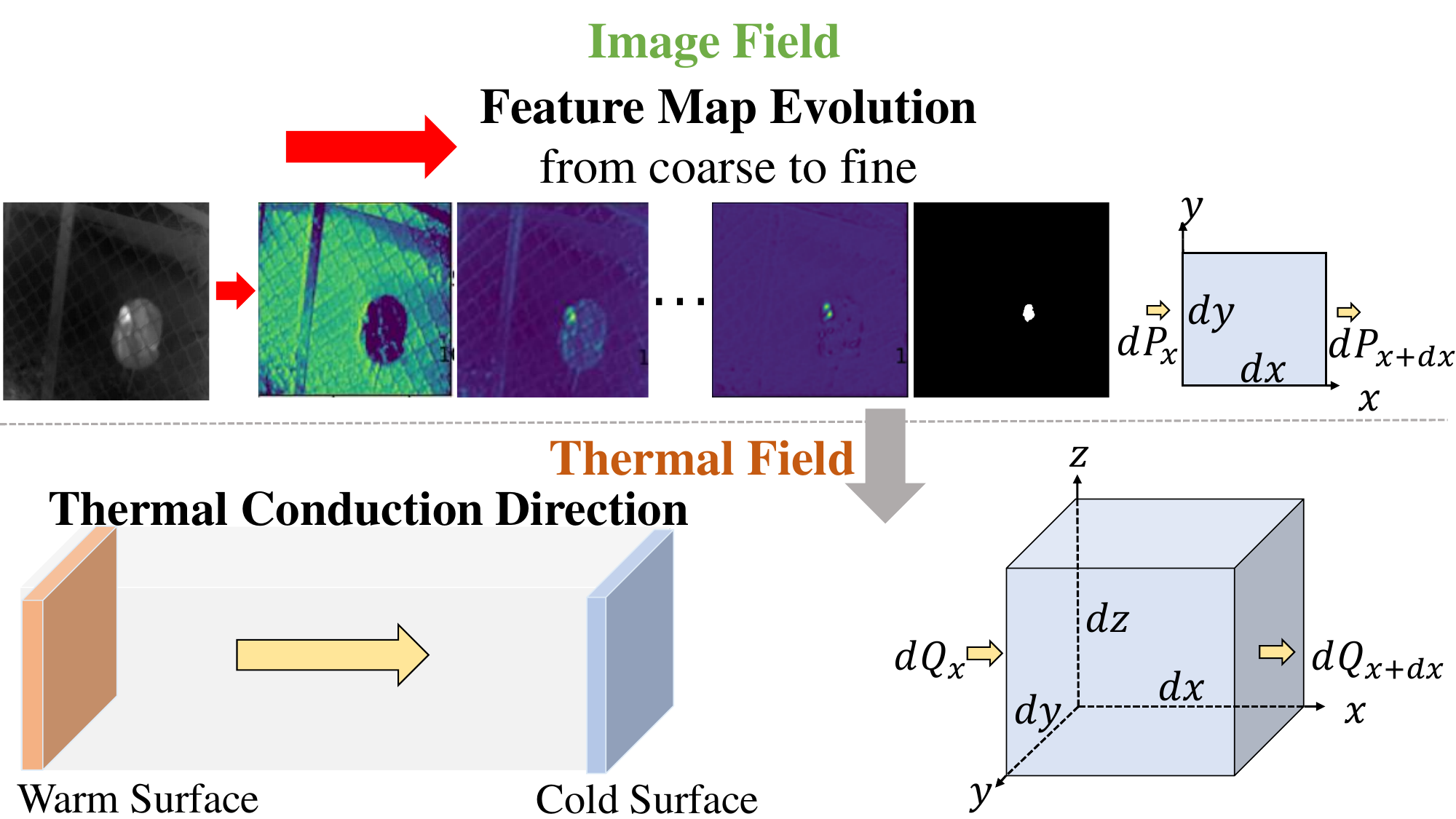}
    \caption{Conversion process between the image ﬁeld and the thermal ﬁeld.The feature map evolution process in the image ﬁeld can be analogous to the thermal conduction process in the thermal ﬁeld. The upper part depicts the image ﬁeld and presents the from-coarse-to-fine feature map evolution in the ISTD process. The upper right corner shows the change of pixel value in a 2-D image micro-element. The lower part shows the thermal conduction process of a 3-D micro-element in the thermal ﬁeld, where thermal energy is conducted spontaneously from high-temperature areas to low-temperature areas.
    }\label{compare}
\end{figure}

Recent years have witnessed the research focus of ISTD shifting to deep-learning-based methods. Deep-learning-based methods improve the ISTD performance by a large margin and can be further classidied into CNN-based methods \cite{chen2023bauenet,dai2021attentional,dai2021asymmetric,zhang2021agpcnet,wang2019miss,li2022dense,zhang2021agpcnet,zhang2022isnet,zhu2023sanet,weng2023ifunet++,du2023bpr} and hybrid methods (methods combining ViT and CNN) \cite{wang2022interior,qi2022ftc,chen2023abmnet,liu2021infrared,zhang2022rkformer,chen2023fluid}. However, despite different module designs, these methods rarely explore a new perspective to look at ISTD, which helps constructing an explainable ISTD network and proposing a potential future research direction. To this end, we propose to understand the feature map evolution process of ISTD from the perspective of thermal conduction.

%Intuitively, we can regard the evolution of the image during ISTD as large numbers of pixels ﬂowing among similar particles in the ﬂuid. The intuitive analogy between ISTD and FD is presented in Fig.~\ref{compare}. Motivated by this analogy, useful theories can be borrowed from ﬂuid dynamics (FD) to improve ISTD performance and promote the research in this area, which is now unexplored. In FD, the fluid elements at the fluid intersecting surface exhibit different density distribution at different time. Similarly, the ISTD process can be perceived as a series of intermediate images that change over time.  

In thermodynamics, micro-elements with different heat exhibit different distribution over time in a closed system. Influenced by the heat source and other external factors, heat will spontaneously be conducted from warm areas to cold areas. Similarly, the ISTD process can be regarded as a series of feature maps that change over time constrained by an objective function. The intuitive analogy between ISTD and thermal dynamics is shown in Fig.~\ref{compare}. The upper part shows the feature map evolution process during ISTD, which is a from-coarse-to-fine process gradually highlighting targets using the adjacent pixel information. Speciﬁcally, in convolution operations, pixels are determined by multiple adjacent pixels of the previous layer. During ISTD process, the micro-elements with different pixel values in the image move under the constraints of an objective function until some micro-elements with high pixel values gather near small target areas. In this way, the small targets gradually get highlighted. The three consecutive images in the upper right part visualize this process. The lower part describes the spontaneous thermal conduction from high-temperature to low-temperature areas. The bottom right image shows the inflow and outflow of the thermal energy in a 3-D micro-element. The two processes are essentially very similar, so some thermodynamic theories can be transferred to ISTD. The most related work of our paper is \cite{zhang2022heat}, which understands super resolution from thermodynamics perspective, but our task, network modules and the way of analogizing thermal field to image field (the pixel movement of ISTD is directed to target areas, but for super resolution it is unordered) are all different.

In this paper, we explore a novel research routine by analogizing the pixel movement during ISTD process as thermal conduction in thermodynamics and propose TCI-Former. Based on the thermal conduction differential equation, we derive the pixel movement differential equation (PMDE) in the image domain for ISTD. Our PMDE builds a spatial–temporal constraint to guide the pixel ﬂow direction, so we design our network based on it. On the one hand, we apply the finite difference method to PMDE and propose thermal conduction-inspired attention (TCIA) to help extracting the main body features of targets. On the other hand, only focusing on main body areas of targets inevitably causes errors in segmenting target boundary areas, so we devise thermal conduction boundary module (TCBM) to refine target body features with fine boundary details.

Our contributions can be summarized in three folds:
\begin{itemize}
    \item We are the first to realize the intrinsic consistency between thermal micro-elements and the image pixels during feature map evolution in ISTD, where the change of heat distribution over time is analogous to the change of pixel values due to pixel movement in consecutive feature map series. We transfer heat conduction theories into the ISTD network design and propose TCI-Former.
    \item Inspired by the thermal conduction differential equation, we derive our pixel movement differential equation (PMDE) to establish a link between spatial and temporal information of pixel values during ISTD process.  
    \item We incorporate the ﬁnite difference method to PMDE and propose thermal conduction-inspired attention (TCIA) to extract target main body features but brings slight errors to target boundary areas. As complement, thermal conduction boundary module (TCBM) is also devised to supplement the target body features with fine boundary details to make up for the errors.
    \item Our method outperforms others on IRSTD-1k and NUAA-SIRST in terms of evaluation metrics.
\end{itemize}

\section{Related Work}
\subsection{Infrared Small Target Detection Networks}

ISTD networks are generally classified into CNN-based and hybrid types. CNN-based networks mainly extract local features. Dai et al. \cite{dai2021asymmetric} released the first public ISTD dataset and proposed asymmetric contextual modulation for cross-layer feature fusion. They then proposed AlcNet \cite{dai2021attentional} to preserve local features of small targets. Wang et al. were the first to apply GAN to ISTD and proposed MDvsFA \cite{wang2019miss}, which achieved a trade-off between missed detection and false alarm. DNANet \cite{li2022dense} devised a dense nested interactive module (DNIM) to progressively interact different level features. ISNet \cite{zhang2022isnet} designed a simple Taylor finite difference-inspired block and a two-orientation attention aggregation module to detect targets.

However, only local features are insufficient to detect all infrared targets because the low contrast background makes many small targets unclear to find. Therefore, researchers turn to hybrid methods \cite{chen2022irstformer,wang2022interior,zhang2022rkformer} by combining ViT with CNN to complement local features with global dependencies. For example, Chen et al. novelly built a ViT-CNN structure based on fluid dynamics for shape-aware ISTD. %IRSTFormer\cite{chen2022irstformer} adopted hierarchical ViT to model long-range dependencies to suppress false alarms but laid insufficient emphasis on local details. IAANet \cite{wang2022interior} concatenates the local patch outputs from a simple CNN structure with a ViT but limits the extraction of global information. RKformer \cite{zhang2022rkformer} introduced the Runge-Kutta method to the encoder block design, which CNN with ViT are coupled.

The above ISTD networks focus on building either feature extraction blocks or fusion modules, none of them provide a new understanding of ISTD from the feature map evolution perspective. In this paper, we open a novel research perspective by abstracting the directional movement of pixels with high pixel values to target areas in the ISTD process as heat conduction from warm to cold areas in thermodynamics.

\subsection{Thermal Conduction Differential Equation}
Thermal conduction studies the law of thermal energy transfer due to temperature difference. Wherever there exists a temperature difference, there is a spontaneous conduction of thermal energy from a high-temperature object to a low-temperature object, or from a high-temperature object part to a low-temperature part \cite{borgnakke2022fundamentals}.

As the basic law of thermal conduction, thermal conduction differential equation indicates that the heat passing through a given section in unit time is proportional to the rate of temperature change and the area of the section perpendicular to the direction of the section. It is the mathematical expression of the differential form of the temperature distribution in the thermal conduction temperature field. The thermal conduction direction is opposite to the temperature increase direction. The equation is established according to the heat conservation law and Fourier law. The law of heat conservation can be expressed as $\Delta Q=\Delta E+ Q_{f}$, where $Q$ is the difference between the thermal energy imported and exported from an object. $\Delta E$ is the increment of internal energy of the object. $Q_{f}$ is the heat of formation of the internal heat source in the object. The Fourier law describes the relationship between thermal conductivity and temperature gradient, which is described as $q=-\lambda \frac{\partial T}{\partial n}$, where $\frac{\partial T}{\partial n}$ is the temperature gradient and $\lambda$ is the thermal conduction coefficient. Rewrite heat conservation equation into the differential form of unit time and space and plug the Fourier Law into the heat conservation equation, we can get the thermal conduction differential equation as follows:
\begin{equation}\label{tcde}
\frac{\partial T}{\partial t}=\frac{\lambda}{\rho c} (\frac{\partial^{2} T}{\partial x^{2}}+\frac{\partial^{2} T}{\partial y^{2}}+\frac{\partial^{2} T}{\partial z^{2}})+\frac{q_{v}}{\rho c},\\
\end{equation}
where $q_{v}$ is the heat of formation of the internal heat source in an object in per unit volume and time.
\begin{figure*}
    \centering
    \includegraphics[width=0.95\textwidth,height=0.48\textwidth]{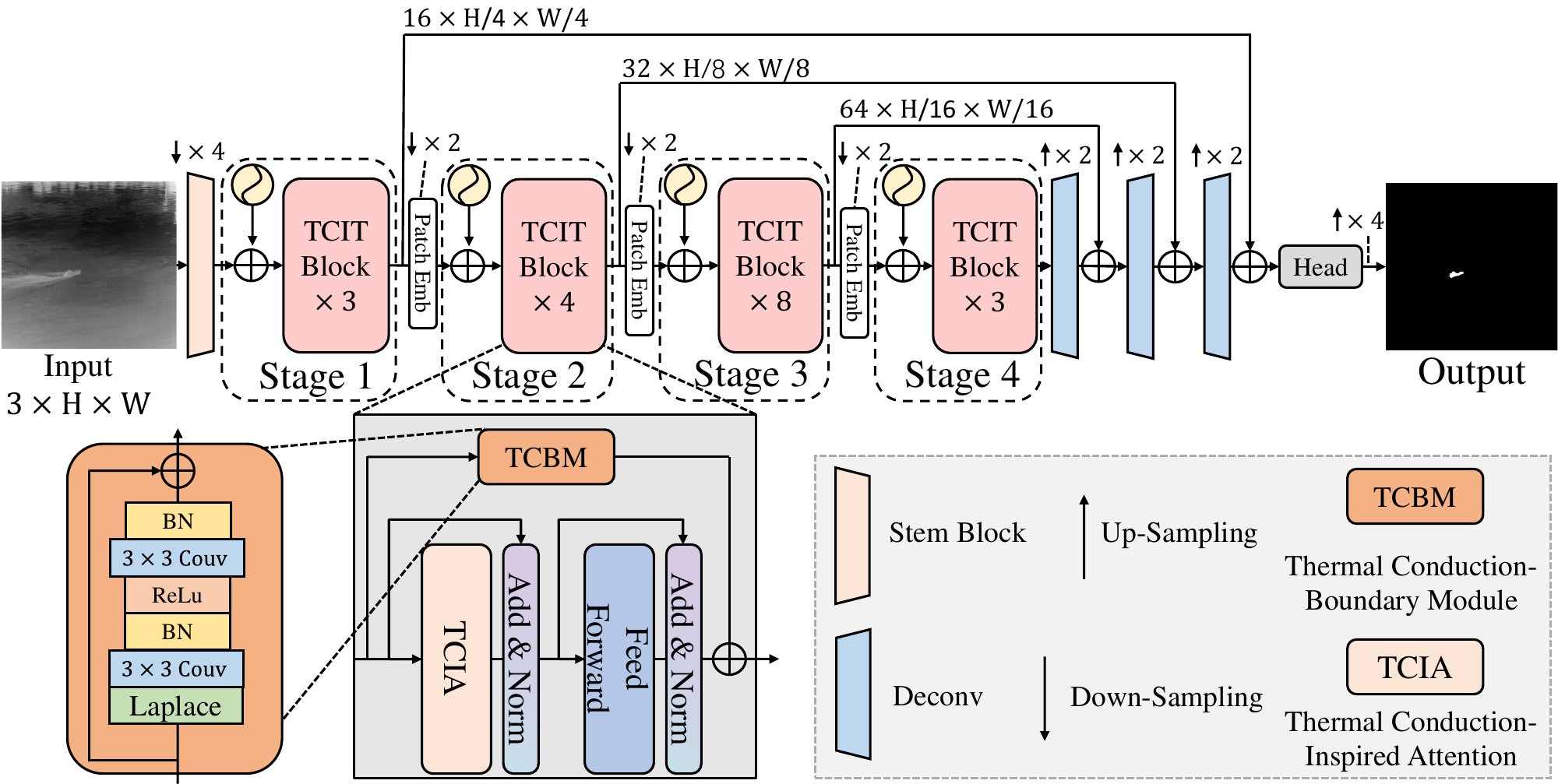}
    \caption{Overall architecture of our TCI-Former with an encoder-decoder structure. The encoder is composed of several TCIT blocks. Each TCIT block contains two key components: Thermal Conduction-Inspired Attention (TCIA) and Thermal Conduction Boundary Module (TCBM), which are both devised based on our derived pixel movement differential equation (PMDE). PMDE is inspired by the thermal conduction differential equation (TCDE) in heat dynamics.}\label{overview}
\end{figure*}

\section{Pixel Movement Differential Equation (PMDE)}
 In a unit of time, the thermal change of a micro-element can be expressed as: [the difference between the imported and exported heat] + [the thermal energy generated by the internal heat source] = [the increase in thermodynamic energy]. The difference between the imported and exported heat corresponds to the feature map pixel value difference between inﬂow and outﬂow ($\Delta P_{f}$). The thermal energy generated by the internal heat source corresponds to the change in the pixel’s own value ($\Delta P_{s}$). The total increase in thermodynamic energy corresponds to the overall change in pixel value ($\Delta P$). Accordingly, in the image ﬁeld we have:
\begin{equation}\label{law}
[\Delta P_{f}]+ [\Delta P_{s}]=[\Delta P].\\
\end{equation}
Similar to the derivation of TCDE, the Pixel Movement Differential Equation (PMDE) can be derived as follows.

\subsection{Pixel Value Difference between Inﬂow and Outﬂow} Within $dt$, we denote the pixel values ﬂowing into the micro-element along the $x$-axis and $y$-axis as $dP_{x}$ and $dP_{y}$, respectively. Similarly, there are also pixel values ﬂowing out of the micro-element along both axes, which we describe as $dP_{x+dx}$ and $dP_{y+dy}$, respectively. Subsequently, according to the relationship between the difference and the derivative, and combine the pixel value difference in the $x$-direction and the pixel value difference in the $y$-direction to get the whole value difference
\begin{equation}
\begin{split}
dP_{x}=p_{x}dydt, dP_{y}=p_{y}dxdt,&\\
dP_{x+dx}=p_{x+dx}dydt=(p_{x}+&\frac{\partial p_{x}}{\partial x}dx)dydt,\\
dP_{y+dy}=p_{y+dy}dxdt=(p_{x}+&\frac{\partial p_{y}}{\partial y}dy)dxdt,\\
\Delta P_{f}=dP_{x+dx}-dP_{x}+&dP_{y+dy}-dP_{y}\\
=-(\frac{\partial p_{x}}{\partial x}+\frac{\partial p_{y}}{\partial y})dxdydt,\\
\end{split}
\end{equation}
where $dp_{x}$, $dp_{y}$, $dp_{x+dx}$, $dp_{y+dy}$ are respectively the inflow and outflow pixel value intensity along the $x$-axis and $y$-axis, which measure the pixel values flowing in and out within per unit area and per unit time. According to the Fourier law in thermodynamics \cite{borgnakke2022fundamentals}, which characterizes the relationship between the heat ﬂow and the micro-element temperature gradient in the heat conduction process, $dp_{x}$, $dp_{y}$ can be calculated as follows:
\begin{equation}\label{fourier}
dp_{x}=-\lambda \frac{\partial P}{\partial x}, dp_{y}=-\lambda \frac{\partial P}{\partial y}.\\
\end{equation}

\subsection{Change in Pixel’s Own Value} For each pixel in the infrared image, its own pixel value changes over time and follows $P_{s}=p_{s}dxdydt$. $p_{s}$ represents pixel intensity, which is the pixel value generated within per unit area and time. $P_{s}$ is the increase of the image micro-element’s pixel value due to its internal points’ spontaneous pixel value changes. Here we only consider the effect of the difference between the imported and exported pixel values, so pixel value of each point is ﬁxed and will not change, which means $p_{s}=0$.

\subsection{Overall Change in Pixel Value} According to the correspondence between the variables in image ﬁeld and heat conduction ﬁeld, we can get the relationship between the pixel value change rate ($\frac{\partial P}{\partial t}$) and the overall pixel value change $\Delta P$ during feature map evolution. The change in the micro-element’s pixel value can be expressed as:
\begin{equation}\label{delta_p}
\Delta P=a\frac{\partial P}{\partial t}dxdydt,\\
\end{equation}
where $a$ is a constant. From Eq.\eqref{law} to Eq.\eqref{delta_p}, we can get the relationship between the pixel value change rate and gradient during the ISTD process, which is the ﬁnal expression of pixel movement differential equation (PMDE):
\begin{equation}\label{pmde}
\frac{\partial P}{\partial t}=\alpha (\frac{\partial^{2} P}{\partial x^{2}}+\frac{\partial^{2} P}{\partial y^{2}}),\\
\end{equation}
where $\alpha=(\lambda/a)$. PMDE builds the link between spatial and temporal information of pixel values in an image. %Some boundary information is contained in the second-order derivative terms because at boundary areas these terms have larger values. 
In the next section we will use the equation to devise two modules which respectively focus on target body and boundary parts to reﬂect the ﬂow of pixels.

\section{Methodology}
\label{sec:method}

\subsection{Overall Architecture}
\label{sec:arc}
The overview of our TCI-Former is displayed in Fig.~\ref{overview}. TCI-Former has a U-Net-like encoder-decoder structure, where the encoder is composed of several Thermal Conduction-Inspired Transformer (TCIT) blocks stacked sequentially while the decoder is built upon three plain deconvolution layers following the common practice. Skip connections are added between the corresponding encoder and decoder layers for cross-layer feature fusion. A fully convolutional segmentation head is connected after the decoder to offer the final predictions. The added circle in the stage blocks denotes the position coding operation for the input tokens. Specifically, each TCIT block contains a Thermal Conduction-Inspired Attention (TCIA) and a Thermal Conduction Boundary Module (TCBM). The TCIT has a parallel structure of global attention and convolution to assemble their merits of modelling local and global information simultaneously. The global attention structure of TCIT block is TCIA, which concentrates on target body information from horizontal and vertical directions in the same way as thermal conduction. The convolutional structure of TCIT is TCBM, which refines target body features with boundary details. %For an infrared input image to our TCI-Former, it is first processed by a stem block, which contains a convolutional layer and a max-pooling layer with a stride of 2 in each layer to downsample the spatial size by $4\times$. Then, the features further go through the encoder-decoder and the segmentation head to get the detected mask.

\subsection{Thermal Conduction-Inspired Attention}

%\begin{figure}
%    \centering
%    \includegraphics[width=0.40\textwidth,height=0.22\textwidth]{surrounding.pdf}
%    \caption{Illustration of pixel value change rule when extracting target body features. The subscripts $i$ and $j$ represent spatial steps, while the superscript $t$ represents the time step, here means the depth of the feature map layer.}\label{surrounding}
%\end{figure}
The finite difference method is a numerical ODE solver. We apply the method to our PMDE to extract the target main body feature, which can be regarded as an approximation of the whole target feature. Thus, we propose TCIA to explore the rule of target body feature extraction during feature map evolution. Here we use the second-order finite difference equation, which is expressed as:
\begin{equation}\label{fdm}
\begin{split}
\frac{\partial^{2} P^{t}_{i,j}}{\partial x^{2}}&=\frac{P^{t}_{i+1,j}-2P^{t}_{i,j}+P^{t}_{i-1,j}}{(\Delta x)^{2}},\\
\frac{\partial^{2} P^{t}_{i,j}}{\partial y^{2}}&=\frac{P^{t}_{i,j+1}-2P^{t}_{i,j}+P^{t}_{i,j-1}}{(\Delta y)^{2}},\\
\end{split}
\end{equation}
where $P^{t}_{i,j}$ is the pixel value in position $(i,j)$ in the $t$-th feature map layer. Applying Eq.\eqref{fdm} to Eq.\eqref{pmde} we have
\begin{equation}\label{pmde2}
\begin{split}
P^{t+1}_{i,j}-P^{t}_{i,j}=\alpha (\frac{P^{t}_{i+1,j}-2P^{t}_{i,j}+P^{t}_{i-1,j}}{(\Delta x)^{2}}+\\
\frac{P^{t}_{i,j+1}-2P^{t}_{i,j}+P^{t}_{i,j-1}}{(\Delta y)^{2}}).\\
\end{split}
\end{equation}
Defining $\Delta x=\Delta y$, we can get the final expression of the target main body part feature extraction rule as follows:
\begin{equation}\label{pmde3}
P^{t+1}_{i,j}=\gamma (P^{t}_{i+1,j}+P^{t}_{i-1,j}+ P^{t}_{i,j+1}+P^{t}_{i,j-1}-4P^{t}_{i,j})+P^{t}_{i,j},\\
\end{equation}
where $\gamma$ denotes $\frac{\alpha}{\Delta x \Delta y}$. Eq.\eqref{pmde3} describes that the pixel value at a certain position in a certain feature map layer is determined by its surrounding pixels in $x$ and $y$ axis of its former layer feature map.

\begin{figure}
    \centering
    \includegraphics[width=0.47\textwidth,height=0.21\textwidth]{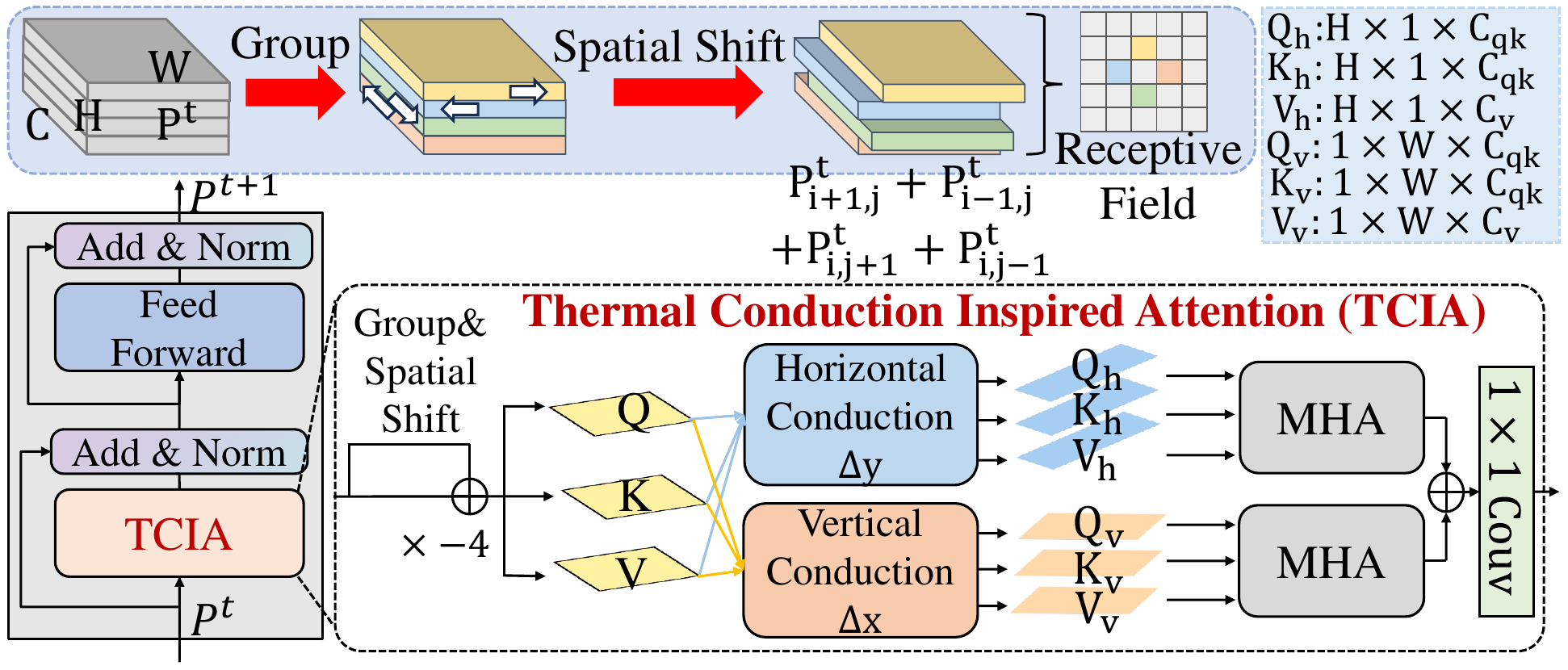}
    \caption{Overall architecture of our proposed Thermal Conduction-Inspired Attention (TCIA), which is devised based on finite difference method and PMDE derived from the TCDE in heat dynamics.}\label{tcia}
\end{figure}

Based on Eq.\eqref{pmde3}, we devise TCIA to extract the main body features of small targets during feature map evolution. Fig.~\ref{tcia} shows the structure of TCIA. The input of TCIA is $P^{t}$ and the output is $\gamma (P^{t}_{i+1,j}+P^{t}_{i-1,j}+ P^{t}_{i,j+1}+P^{t}_{i,j-1}-4P^{t}_{i,j})$, which is obtained through horizontal and vertical conduction attentions $\Delta y$ and $\Delta x$ to aggregate surrounding pixel information of the former layer before element-wise addition with $P^{t}$. The channel of $P^{t}\in \mathbb{R}^{C \times H \times W}$ is divided into four groups before shifting each channel group to different directions by $+1$ or $-1$. In this way, the receptive field of $P^{t}$ after spatial shift is rhombic, which corresponds to $P^{t}_{i+1,j}+P^{t}_{i-1,j}+ P^{t}_{i,j+1}+P^{t}_{i,j-1}$. We can get $P^{t}_{i+1,j}+P^{t}_{i-1,j}+ P^{t}_{i,j+1}+P^{t}_{i,j-1}-4P^{t}_{i,j}$ through residual operation and then linearly project this term into $Q$, $K$, $V$. The horizontal conduction $\Delta y$ is implemented by taking average of query feature map on the horizontal direction. In the same way, the vertical conduction $\Delta x$ squeezes query feature map on the vertical direction. The same operations are also conducted upon $K$ and $V$, so we can get $Q_{h}, K_{h} \in \mathbb{R}^{H\times C_{qk}}$, $V_{h} \in \mathbb{R}^{H\times C_{v}}$ and $Q_{v}, K_{v} \in \mathbb{R}^{W\times C_{qk}}$, $V_{v} \in \mathbb{R}^{W\times C_{v}}$. Each of the two conduction attentions reserves the global information to a single axis, so that each position on the feature map propagates information only on two squeezed $x$-axis and $y$-axis features. Then the $Q,K,V$ vectors are fed into multi-head attentions and then added together for horizontal and vertical feature aggregation to realize the $\gamma (P^{t}_{i+1,j}+P^{t}_{i-1,j}+ P^{t}_{i,j+1}+P^{t}_{i,j-1}-4P^{t}_{i,j})$ term. For the last term $P^{t}_{i,j}$ in Eq.\eqref{pmde3}, it is added through the residual and layer norm operation in the transformer block. In this way, the TCIA based on Eq.\eqref{pmde3} is realized.

\subsection{Thermal Conduction Boundary Module}
TCIA helps extracting target body features, but the features extracted by TCIA branch alone are not fine enough in near boundary regions because the finite difference method used in TCIA is a numerical method, which inevitably brings small errors. A certain degree of dispersion exists due to numerical uncertainty during pixel value movement. To solve this, we need to refine the coarse target body features with fine boundary details to make up for the uncertain errors. We notice that our PMDE itself has already contained boundary information (second-order derivative terms), so to extract target boundary features we design Thermal Conduction Boundary Module (TCBM) based on PMDE. The differential form of Eq.\eqref{pmde} can be described as:
\begin{equation}\label{tcrb1}
P^{t+\Delta t}-P^{t}=\Delta t\alpha (\frac{\partial^{2} P}{\partial x^{2}}+\frac{\partial^{2} P}{\partial y^{2}}.)\\
\end{equation}

During ISTD, the extracted feature maps are arranged in a chronological order. The PMDE establishes the relationship between the change of pixel value in temporal domain ($P^{t+\Delta t}-P^{t}$) and 2-D spatial domain ($\frac{\partial^{2} P^{t}}{\partial x^{2}}$, $\frac{\partial^{2} P^{t}}{\partial y^{2}}$) during feature extraction. Defining the time step $\Delta t$ as 1, we can explore the boundary feature evolution rule between two consecutive feature maps. The specific expression of PMDE can be rewritten as:
\begin{equation}\label{tcrb2}
P^{t+1}-P^{t}=h\alpha (\frac{\partial^{2} P^{t}}{\partial x^{2}}+\frac{\partial^{2} P^{t}}{\partial y^{2}}),\\
\end{equation}
where $t$ means the $t$-th residual calculation. $h$ is the step size between the $t$-th and $t+1$-th residual calculation. The TCBM applies spatial information to make up for the lack of boundary refinements during feature extraction in the encoder. The right side of Eq.\eqref{tcrb2} is the second derivative of $P^{t}$ in the $x$- and $y$-directions, respectively. Thus, with this item, we obtain the spatial information which can be used as the residual supplementary for time information, that is, the information in the forward extraction process. $\frac{\partial^{2} P^{t}}{\partial x^{2}}$ and $\frac{\partial^{2} P^{t}}{\partial y^{2}}$ have larger value at boundary areas, therefore TCBM is sensitive to target boundaries and can play a complementary role to the target body features. Our TCBM incorporates a Laplace operator into a residual block, where the Laplace operator is used to realize the $\frac{\partial^{2} P^{t}}{\partial x^{2}}$ and $\frac{\partial^{2} P^{t}}{\partial y^{2}}$ terms.

\subsection{Loss Function}
Dice loss \cite{sudre2017generalised} measures the difference between a mask prediction and the ground truth. It can also relieve sample imbalance problem and is defined as: 
\begin{equation}\label{loss1}
L_{dice}= 1-\frac{2|X \cap Y|}{|X|+|Y|},
\end{equation}
where $X$ denotes the mask prediction and $Y$ is the ground truth. Our final loss function $L_{Final}$ includes $L_{Seg}$ as the main loss function and Target Boundary loss ($L_{TB}$) and Interior Body loss ($L_{IB}$) as two auxiliary loss functions. $L_{Final}$ is calculated as:
\begin{equation}\label{totalloss}
L_{Final}=L_{Seg}^{hyb}+ L_{TB}^{hyb}+ L_{IB}^{hyb}.\\
\end{equation}
$L_{IB}$ and $L_{Seg}$ share the same $Y$ as the ground truth mask, while the $X$ of $L_{IB}$ is the segmentation head output from the TCIA encoder branch, and the $X$ of $L_{Seg}$ is the final prediction output. The $X$ of $L_{TB}$ is the segmentation head output from the TCBM encoder branch, and the $Y$ of $L_{TB}$ is the boundary mask label.

\section{Experiments}
\subsection{Experimental Settings}
\subsubsection{Datasets.}
We choose NUAA-SIRST \cite{dai2021asymmetric} and IRSTD-1k \cite{zhang2022isnet} as our experimental datasets. NUAA-SIRST contains 427 infrared images of various sizes while IRSTD-1k consists of 1,000 real infrared images of $512\times512$ in size. IRSTD-1k is a more difficult ISTD dataset with richer scenarios. For each dataset, we use 80$\%$ of images as training set and 20$\%$ as test set.
%For each dataset, we use 50$\%$ of images as training set, 30$\%$ as validation set, and 20$\%$ as test set.

%\subsubsection{Compared methods}
%We select some most well-performed ISTD methods for comparison, including traditional methods, CNN-based methods and hybrid methods. Among them, Tophat \cite{rivest1996detection}, PSTNN \cite{zhang2019infrared} and MSLSTIPT \cite{sun2020infrared} are traditional methods; MDvsFA \cite{wang2019miss}, ACM \cite{dai2021asymmetric}, AlcNet \cite{dai2021attentional}, DNANet \cite{li2022dense}, Dim2Clear \cite{zhang2023dim2clear} and ISNet \cite{zhang2022isnet} are CNN-based methods; IAANet \cite{wang2022interior} and RKformer \cite{zhang2022rkformer} are hybrid methods.

\subsubsection{Evaluation Metrics.}
We compare our TCI-Former with other SOTA methods in terms of both pixel-level and object-level evaluation metrics. The pixel-level metrics include Intersection over Union ($IoU$) and Normalized Intersection over Union ($nIoU$), while the object-level metrics include Probability of Detection ($P_{d}$) and False-Alarm Rate ($F_{a}$).

$IoU$ measures the accuracy of detecting the accuracy of detecting the corresponding
object in a given dataset. $nIoU$ is the normalization of $IoU$, which can make a better balance between structural similarity and pixel accuracy of infrared small targets. $IoU$ and $nIoU$ are defined as:
\begin{equation}
IoU=\frac{A_{i}}{A_{u}},nIoU=\frac{1}{N}\sum_{i=1}^{N}(\frac{TP[i]}{T[i]+P[i]-TP[i]}),\\
\end{equation}
where $A_{i}$ and $A_{u}$ are the areas of intersection region and union region between the prediction and ground truth, respectively. $N$ is the total number of samples, $TP[.]$ is the number of true positive pixels, $T[.]$ and $P[.]$ is the number of ground truth and predicted positive pixels. 
 
 $P_{d}$ calculates the ratio of the number of correctly predicted targets $N_{pred}$ to all targets $N_{all}$. $F_{a}$ refers to the ratio of falsely predicted target pixels $N_{false}$ to all the pixels in the infrared image $N_{all}$. $P_{d}$ and $F_{a}$ are calculated as follows:
\begin{equation}
P_{d}=\frac{N_{pred}}{N_{all}},F_{a}=\frac{N_{false}}{N_{all}}.\\
\end{equation}

\subsubsection{Optimization.}
 The algorithm is implemented in Pytorch, with Adaptive Gradient (AdaGrad) as the optimizer with the initial learning rate set to 0.05 and weight decay coefficient set to 0.0004. A Titan XP GPU is used for training, with batch size set to 4. Training on SIRST and IRSTD-1k takes 800 epochs and 600 epochs respectively.
\begin{figure}
    \centering
    \includegraphics[width=0.47\textwidth,height=0.27\textwidth]{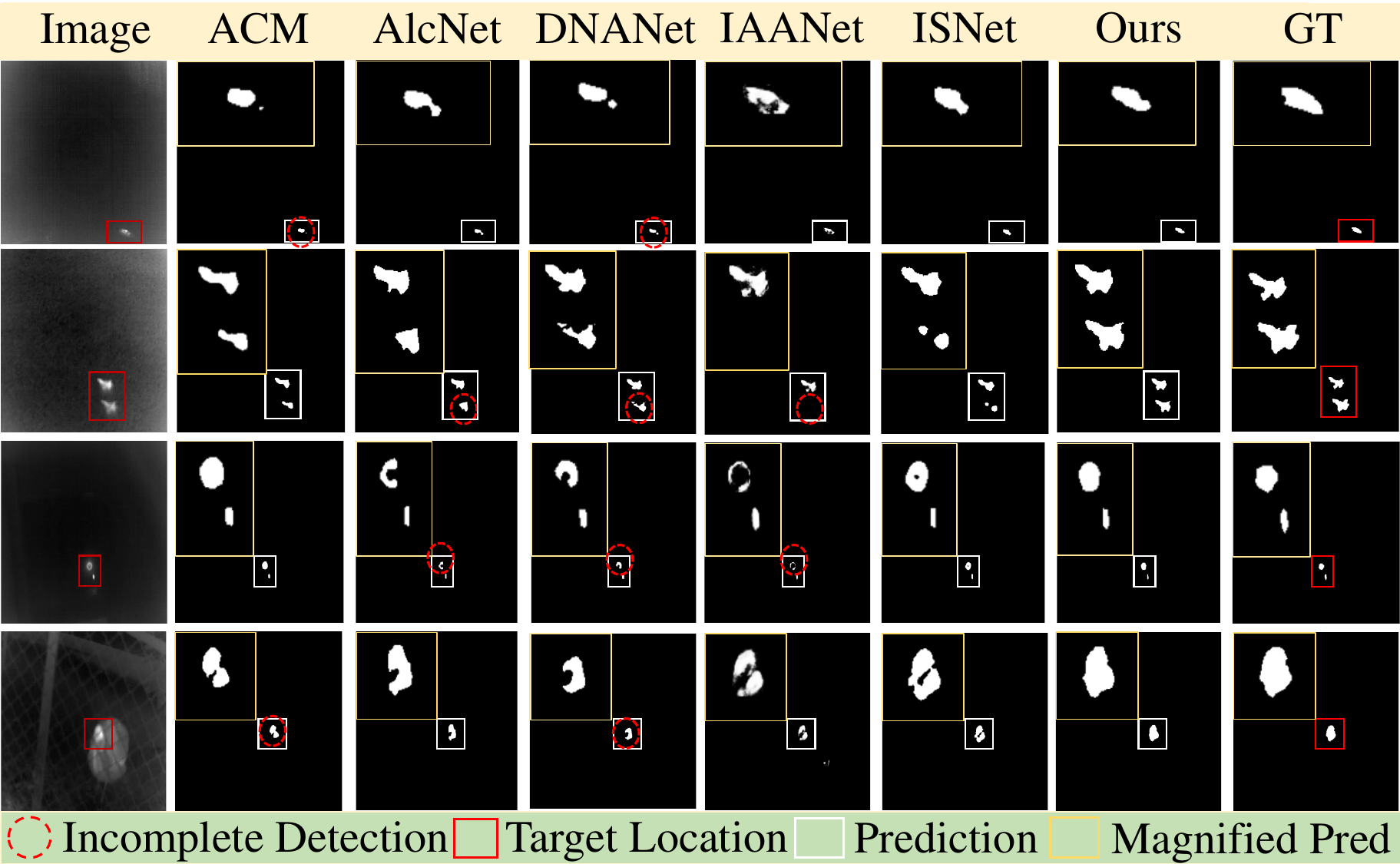}
    \caption{Result visualization of different ISTD methods.}\label{vis}
\end{figure}
\begin{figure}
    \centering
    \includegraphics[width=0.47\textwidth,height=0.25\textwidth]{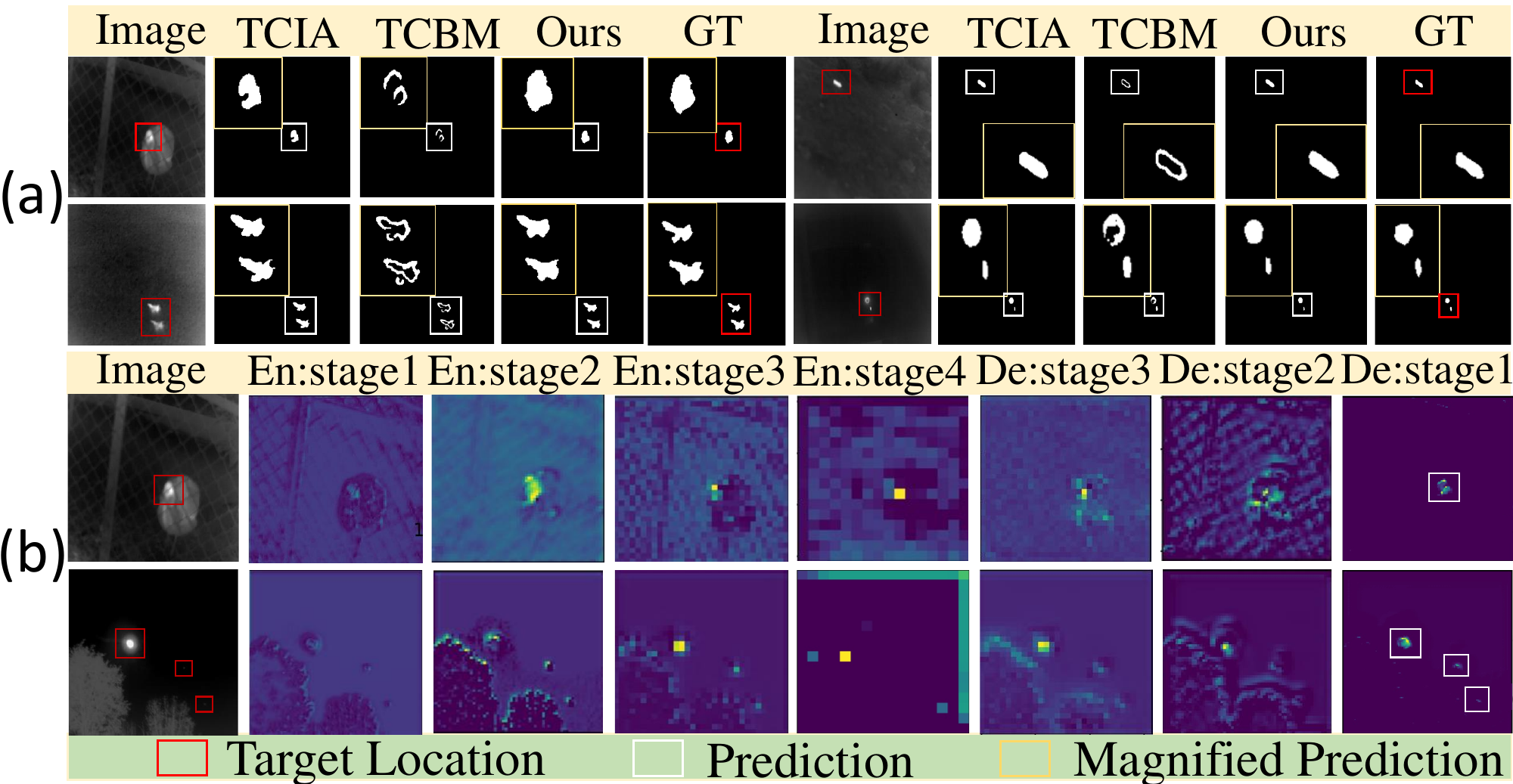}
    \caption{Visualization of (a) TCIA and TCBM branch outputs and (b) intermediate stage feature map evolution.}\label{mid_vis}
\end{figure}

\begin{table*}
\centering
    \begin{tabular}{c|c|cccccccc}
    %\toprule
    \toprule[1pt]
        \multirow{2}{*}{Method} & \multirow{2}{*}{Type}&\multicolumn{4}{c}{NUAA-SIRST} & \multicolumn{4}{c}{IRSTD-1k}\\
        & &IoU $\uparrow$ & nIoU $\uparrow$ &Pd $\uparrow$ &Fa $\downarrow$ & IoU $\uparrow$ &  nIoU $\uparrow$ &Pd $\uparrow$ &Fa $\downarrow$\\
        \midrule
        %\hline
        %Tophat \cite{rivest1996detection}  & Trad& 7.140   & 5.20  & 79.84 & 10.12 & 10.06   & 7.44 & 75.11& 1432\\
        PSTNN \cite{zhang2019infrared}  & Trad & 22.40   & 22.35 &77.95 & 29.11  & 24.57   & 17.93 & 71.99 & 35.26\\
        MSLSTIPT \cite{sun2020infrared}  & Trad & 10.30   & 9.58  & 82.13 & 1131 & 11.43   & 5.93 & 79.03 & 1524\\
        MDvsFA \cite{wang2019miss}  & CNN & 60.30   & 58.26  &89.35 & 56.35 & 49.50   & 47.41 &82.11 &80.33\\
        ACM \cite{dai2021asymmetric}  & CNN & 72.33   & 71.43 &96.33 & 9.325 & 60.97   & 58.02 & 90.58 & 21.78\\
        AlcNet \cite{dai2021attentional} & CNN & 74.31   & 73.12 & 97.34 & 20.21 & 62.05   & 59.58 & 92.19 & 31.56\\
        DNANet \cite{li2022dense} & CNN & 75.27   & 73.68 & 98.17& 13.62 & \underline{69.01}   & \underline{66.22} & 91.92 & 17.57\\
        Dim2Clear \cite{zhang2023dim2clear}  & CNN& 77.20   & 75.20 &99.10 & 6.72  & 66.3   & 64.2 & 93.7& 20.9\\
        FC3-Net \cite{zhang2022exploring} & CNN& 74.22   & 72.64 &99.12 & 6.569  & 64.98   & 63.59 & 92.93& 15.73\\
        IAANet \cite{wang2022interior}  & Hybrid& 75.31   & 74.65 & 98.22&  35.65 & 59.82   & 58.24 & 88.62& 24.79\\
        RKformer \cite{zhang2022rkformer} & Hybrid&77.24   & 74.89  & 99.11& \textbf{1.580}  & 64.12  & 64.18 &93.27 & 18.65\\
        ISNet \cite{zhang2022isnet} & CNN& \underline{80.02}  & \underline{78.12}  & \underline{99.18} &4.924 & 68.77   & 64.84 & \underline{95.56}& \underline{15.39}\\
        TCI-Former & Hybrid& \textbf{80.79} & \textbf{79.85}&\textbf{99.23} &\underline{4.189} & \textbf{70.14} & \textbf{67.69}& \textbf{96.31} & \textbf{14.81}\\
        %\bottomrule
        \bottomrule[1pt]
    \end{tabular}
    \caption{Quantitative  results  of  different  methods on NUAA-SIRST and IRSTD-1k.  The figures in bold and underline mark the highest and the second highest ones in each column.} \label{table1}
\end{table*}

\subsection{Comparison with SOTA Methods}

\subsubsection{Quantitative Comparisons.}
We select some SOTA ISTD methods for comparison. As shown in Table~\ref{table1}, our TCI-Former performs the best in terms of pixel-level and object-level metrics on both datasets.

For the pixel-level metrics ($IoU, nIoU$), the deep-learning methods generally surpass the traditional methods because deep-learning methods do not rely heavily on prior knowledge and handcraft features as traditional methods do. However, deep-learning methods lay insufficient emphasis on target edges, causing limited $IoU$ and $nIoU$. %Notably, the value of the $IoU$ index is often higher than the $nIoU$ index for all methods. This is because the $IoU$ index pays more attention to small targets, while the $nIoU$ index focuses more on larger targets. 
Our TCI-Former achieves the best performance on both $IoU$ and $nIoU$, meaning that our method achieves the best shape-aware segmentation performance thanks to our TCBM. 

For the object-level metrics ($P_{d}, F_{a}$), how to reach a trade-off between $P_{d}$ and $F_{a}$ is challenging because the two metrics are mutually exclusive. Traditional methods fail to balance the two metrics but deep-learning methods make it. Our TCI-Former achieves the best object-level metrics results except that our $F_{a}$ is second only to RKformer \cite{zhang2022rkformer} in NUAA-SIRST. However, our $F_{a}$ significantly outperforms it in IRSTD-1k, which is a more difficult ISTD dataset with richer scenarios. The results demonstrate that our method can learn better representations to find the small targets covered by low contrast and noisy background owing to our TCIA, which mimics thermal conduction to extract target main body features.

\subsubsection{Visual Comparisons.}
Visual results with closed-up views of different methods is shown in Fig.~\ref{vis}. As shown in Fig.~\ref{vis}, most CNN-based methods suffer incomplete detection for lack of extracting global contexts. Hybrid method generally outperforms CNN-based methods with fewer severely incomplete detection cases, but still cannot predict accurate target shapes. Compared with other methods, our method significantly curtails bad cases and achieves better boundary-aware segmentation of small targets. This is because our network can not only extract target body features like thermal conduction, but also refine body features with fine boundary information. %In this way, small targets can get precisely located and their shapes are recognized. 

To demonstrate the target body location effect of TCIA and boundary refinement effect of TCBM, we visualize the segmentation head outputs of TCIA branch and TCBM branch in Fig.~\ref{mid_vis} (a). To present the from-coarse-to-fine feature map evolution process, we visualize intermediate feature maps of all stages in encoder (En: stage1,2,3,4) and decoder (De: stage3,2,1) in Fig.~\ref{mid_vis} (b). We can find that the small target areas gradually get highlighted like heat conducted from warm to cold areas from the decoder stage 3,2,1 feature maps, which complies with our analogy.

\subsection{Ablation Study}
\subsubsection{Impact of Each Module.}
The ablation study of TCIA and TCBM is shown in Table~\ref{ablation1}. The baseline uses basic pyramid ViT \cite{wang2022pvt} as encoder. Table~\ref{ablation1} demonstrates the positive effects of both designs and combining them together brings the best results, implying that they are complementary to each other. The reason is that ViT block equipped with TCIA can extract main target body features from surrounding areas in orthogonal directions, while TCBM in parallel refines the coarse body features with boundary details to improve detection performance.
\begin{table}
\centering
    \begin{tabular}{c|cccc}
    \toprule[1pt]
        Method &  IoU $\uparrow$& nIoU $\uparrow$& Pd $\uparrow$&  Fa $\downarrow$\\
        \midrule
        %\hline
        Baseline   & 62.82   & 60.59   &  92.97  & 26.37\\
        +TCIA   & 67.26   & 65.03   &  94.55  & 19.83\\
        +TCIA+TCBM &\textbf{70.14} & \textbf{67.69}& \textbf{96.31} & \textbf{14.81}\\
        \bottomrule[1pt]
    \end{tabular}
    \caption{Ablation study of each module on IRSTD-1k.} \label{ablation1}
\end{table}
\subsubsection{Impact of TCIA.} 
To ablate TCIA, we compare our TCIA with multi-head self-attention (MHSA) \cite{wang2022pvt}, cross-shaped window self-attention (CSWSA) \cite{dong2022cswin} and the multi-head relation attention (MHRA) \cite{li2022uniformer}. As shown in Table~\ref{ablation2}, our TCIA outperforms others in all metrics, showing better small target location ability. The reason is that in TCIA the spatial shift operation enables the encoder block to be more aware of boundaries, which helps extracting more complete target body features. The superiority of TCIA demonstrates our analogy between ISTD process and thermal conduction process is effective. 
\begin{table}
\centering
    \begin{tabular}{c|cccc}
    \toprule[1pt]
        Method &  IoU $\uparrow$& nIoU $\uparrow$& Pd $\uparrow$&  Fa $\downarrow$\\
        \midrule
        %\hline
        MHSA  & 66.73   & 64.69   &  94.05  & 19.22\\
        CSWSA  & 68.23   & 66.05   &  95.36  & 17.41\\
        MHRA  & 68.86   & 66.87   & 95.74   & 16.76\\
        TCIA & \textbf{70.14} & \textbf{67.69}& \textbf{96.31} & \textbf{14.81}\\
        \bottomrule[1pt]
    \end{tabular}
    \caption{Ablation study of TCIA on IRSTD-1k.} \label{ablation2}
\end{table}

\subsubsection{Impact of TCBM.} 
In Table~\ref{ablation3} we compare TCBM (Laplace+Resblock) with basic Resblock and basic Resblock with Roberts operator to examine the boundary feature extraction effect of different designs. Our TCBM delivers the best result, because (1) edge operators help basic Resblock to extract edges and (2) the edges extracted by Roberts operator is thick and less accurate. 
\begin{table}
\centering
    \begin{tabular}{c|cccc}
    \toprule[1pt]
        Method &  IoU $\uparrow$& nIoU $\uparrow$& Pd $\uparrow$&  Fa $\downarrow$\\
        \midrule
        %\hline
        ResBlock  & 68.93   & 66.62   & 95.90   & 16.35\\
        Roberts+ResBlock  & 69.51   & 67.38   &  96.02  & 15.70\\
        TCBM & \textbf{70.14} & \textbf{67.69}& \textbf{96.31} & \textbf{14.81}\\
        \bottomrule[1pt]
    \end{tabular}
    \caption{Ablation study of TCBM on IRSTD-1k.} \label{ablation3}
\end{table}

\subsection{Model Complexity Analysis}
We also compare the model complexity of different methods in terms of parameter number (M), FLOPs (G) and inference time (s), as shown in Table~\ref{table_complex}. Compared with other methods, our method doesn’t have many parameters and has acceptable FLOPs and inference time. This is because we squeeze the dimensions of $q, k, v$ before attention operations in our TCIA, which reduces model parameters and improves efficiency. Our model reach a general balance among different model complexity indicators.

\begin{table}[!t]
\centering
    \begin{tabular}{c|ccc}
    \toprule[1pt]
        Method & Param & FLOPs & Inf\\
        %\cline{2-9}
        \midrule
        %\hline
        %MDvsFA \cite{wang2019miss}  & 6.28& 1644.7&0.24\\
        ACM \cite{dai2021asymmetric}  & 0.52& 2.02&0.01\\
        DNANet \cite{li2022dense} & 4.7 & 56.34 &0.15\\
        IAANet \cite{wang2022interior}  & 14.05 & 18.13&0.29\\
        RKformer \cite{zhang2022rkformer} & 29.00 & 24.73&0.08\\
        %ISNet \cite{zhang2022isnet} & 1.09& 122.55&0.05\\
        TCI-Former & 3.66 & 5.87 & 0.04\\
        \bottomrule[1pt]
    \end{tabular}
    \caption{Comparison of the model parameters (M), FLOPs (G) and inference time (s) of different methods.} \label{table_complex}
\end{table}

\section{Conclusion}
      Motivated by the analogy of pixel movement during ISTD process and thermal conduction in thermodynamics, we propose TCI-Former for ISTD. We first derive PMDE for the image domain from thermodynamic equation. We then apply finite difference method to PMDE and devise TCIA and embed it into encoder block to extract target main body features by simulating the thermal conduction process. We also propose TCBM based on PMDE to parallelly refine the target body features with fine boundary details. Experiments on NUAA-SIRST and IRSTD-1k prove the superiority of TCI-Former, which explores a new research routine.
      
\section{Acknowledgments}
This work is supported by the National Natural Science Foundation of China (No.62121002, No. 62272430, No. U20B2047) and the Fundamental Research Funds for the Central Universities.

\bibliography{aaai24}

\end{document}